\documentclass[conference]{IEEEtran}
\IEEEoverridecommandlockouts
\usepackage{cite}
\usepackage{amsmath,amssymb,amsfonts}
\usepackage{algorithmic}
\usepackage{graphicx}
\usepackage{textcomp}
\usepackage{xcolor}
\usepackage{multirow}

\def\BibTeX{{\rm B\kern-.05em{\sc i\kern-.025em b}\kern-.08em
    T\kern-.1667em\lower.7ex\hbox{E}\kern-.125emX}}
\begin{document}

\title{Optimized Design Method for Satellite \\Constellation Configuration 
Based on Real-time Coverage Area Evaluation\\

}

\author{\IEEEauthorblockN{Jiahao Zhou\textsuperscript{1}, Boheng Li\textsuperscript{2}, Qingxiang Meng\textsuperscript{1*}}
\IEEEauthorblockA{{\textsuperscript{1}School of Remote Sensing and Information Engineering, Wuhan University, Wuhan, China} \\
{\textsuperscript{2}School of Cyber Science and Engineering, Wuhan University, Wuhan, China}\\
\textsuperscript{*}Corresponding author, e-mail: mqx@whu.edu.cn \\
}

}

\maketitle

\begin{abstract}
When using constellation synergy to image large areas for reconnaissance, it is required to achieve the coverage capability requirements with minimal consumption of observation resources to obtain the most optimal constellation observation scheme. With the minimum number of satellites and meeting the real-time ground coverage requirements as the optimization objectives, this paper proposes an optimized design of satellite constellation configuration for full coverage of large-scale regional imaging by using an improved simulated annealing algorithm combined with the real-time coverage evaluation method of hexagonal discretization. The algorithm can adapt to experimental conditions, has good efficiency, and can meet industrial accuracy requirements. The effectiveness and adaptability of the algorithm are tested in simulation applications.
\end{abstract}

\vspace{1ex}

\section{Introduction}
Satellite ground-imaging coverage technology is widely used [1], and research on this technology continuously intensifies. In constellation optimization design, scenarios are often encountered in which multiple imaging satellites need to be used to collaborate in imaging observations of a larger regional target. To ensure satellite coverage of a specific target or region while reducing the cost required to complete the mission, it is essential to design a reasonable constellation distribution of satellites for ground imaging coverage.

\vspace{1ex}
Consider the following satellite usage scenario: within a short time zone, the user department urgently needs image data for a certain larger area, but due to the short time given and the small number of coverage opportunities, the area cannot be captured in its entirety even if all coverage opportunities are used [2]. In this case, to maximize the coverage efficiency as much as possible, it is desired to develop a constellation-optimized design solution that achieves the required average coverage of the target area with the least consumption of resources. The problem is optimizing satellite constellation configuration based on coverage area evaluation in the resource-limited scenario.

\vspace{1ex}
In this paper, we develop a novel solution model based on the hexagonal discretization technique and simulated annealing algorithm, and propose a new solution strategy to optimize the constellation configuration that meets the maximum observation area requirement, and output a variety of different satellite constellation configurations for selection. Simulation experiments show that the proposed strategy can obtain high-quality solutions in an acceptable time.

\section{Satellite orbit coverage model}

\subsection{Calculation of subsatellite point track}
During the operation of remote sensing satellites, the subsatellite tracks of their adjacent orbital periods cannot be completely coincident due to the influence of the Earth's rotation and ingress, which is one of the most basic bases for designing the return orbit. Therefore, to calculate the subsatellite track during the operation of the satellite, the results of the subsatellite track in a single orbital period need to be calculated first.

\vspace{1ex}
The track of the subsatellite point is usually expressed by the right ascension $\lambda$ and declination $\varphi$. When only the effect of Earth's rotation is considered[3], we can obtain the right ascension and declination of the satellite's hypostasis directly according to the six roots of the satellite's orbit shown in Table 1, and the position parameters of the satellite's hypostasis on Earth are shown in Fig. 1.

\begin{figure}[htbp]
\centerline{\includegraphics[width=1.7in]{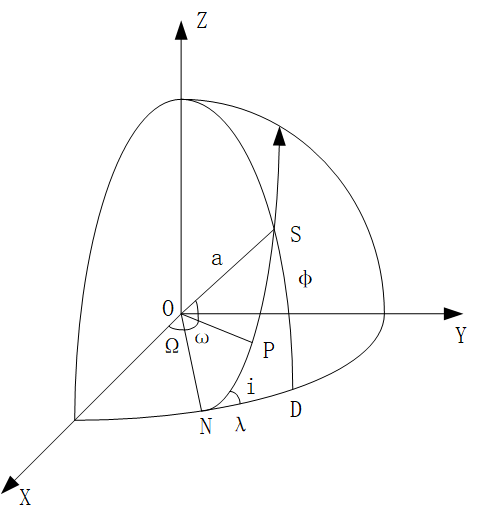}}
\caption{Illustration of the parameters of the subsatellite point position}
\label{fig}
\end{figure}

\begin{table}[htbp]
\caption{Significance of the number of orbital elements}
\centering
\begin{tabular}{|c|c|}
\hline
Symbol & Descripton    \\ \hline
$a$  & semi major axis   \\ \hline
$e$ & eccentricity   \\ \hline
$i$ & inclination  \\ \hline
$\Omega$  & longitude of ascending node \\ \hline
$\omega$  & argument of periapsis   \\ \hline
$M$  & true anomal\\ \hline
\end{tabular}
\end{table}

For the calculation of multiple orbital periods,the influence of the Earth's uptake also needs to be considered. In this paper, we adopt the $J2$ uptake model commonly used to fit satellite orbits, at which time, let $\omega_{E}$ be the angular velocity of the Earth's rotation, the satellite's subsatellite point track is calculated as

\begin{small}
\begin{equation}
\lambda = \Omega_{0} + {\mathit{\arctan}( {{\mathit{\cos}(i)}\cdot{\mathit{\tan}({\omega + f}))}} - S_{G0} - \left( {\omega_{E} - \frac{d\Omega}{dt}} \right)\left( {t - t_{0}} \right)\label{eq}}
\end{equation}
\end{small}

\begin{equation}
\varphi = {\mathit{\arcsin}({{\mathit{\sin}(i)}\cdot{\mathit{\sin}( {\omega + f} )}} )}\label{eq}
\end{equation}

For the average root, a, e, and i do not change and $\Omega$, $\omega$, and M change with time.

\begin{equation}
\frac{d\Omega}{dt} = - \frac{3nJ_{2}R_{e}^{2}}{2a^{2}\left( {1 - e^{2}} \right)^{2}}{\mathit{\cos}(i)}\label{eq}
\end{equation}

\begin{equation}
\frac{d\omega}{dt} = - \frac{3nJ_{2}R_{e}^{2}}{2a^{2}\left( {1 - e^{2}} \right)^{2}}\left( \frac{5}{2}{sin}^{2}(i) - 2 \right)\label{eq}
\end{equation}

\begin{equation}
\frac{dM}{dt} = n - \frac{3nJ_{2}}{2\sqrt{\left( {1 - e^{2}} \right)^{3}}}\left( \frac{R_{e}}{a} \right)^{2}\left( \frac{3}{2}{sin}^{2}(i) - 1 \right)\label{eq}
\end{equation}

The track of the subsatellite points of adjacent orbital periods are the same, and their longitude differences are

\begin{equation}
\Delta\lambda_{2\pi} = - \left( {\omega_{E} - \frac{d\Omega}{dt}} \right)T\label{eq}
\end{equation}

Here, $J_{2}$ is the uptake coefficient, while $n$ is the velocity of the earth's rotation in translational motion

\vspace{1ex}
Therefore, if the six orbital roots at the moment $t_0$ and the Greenwich sidereal time $S_G (t_0)$ are known, we can perform the simulation calculation of the subsatellite point track. The subsatellite point track calculated according to equation (1)(2)(3)(4)(5)(6) is shown in Fig. 2.

\begin{figure}[htbp]
\centerline{\includegraphics[width=2.8in]{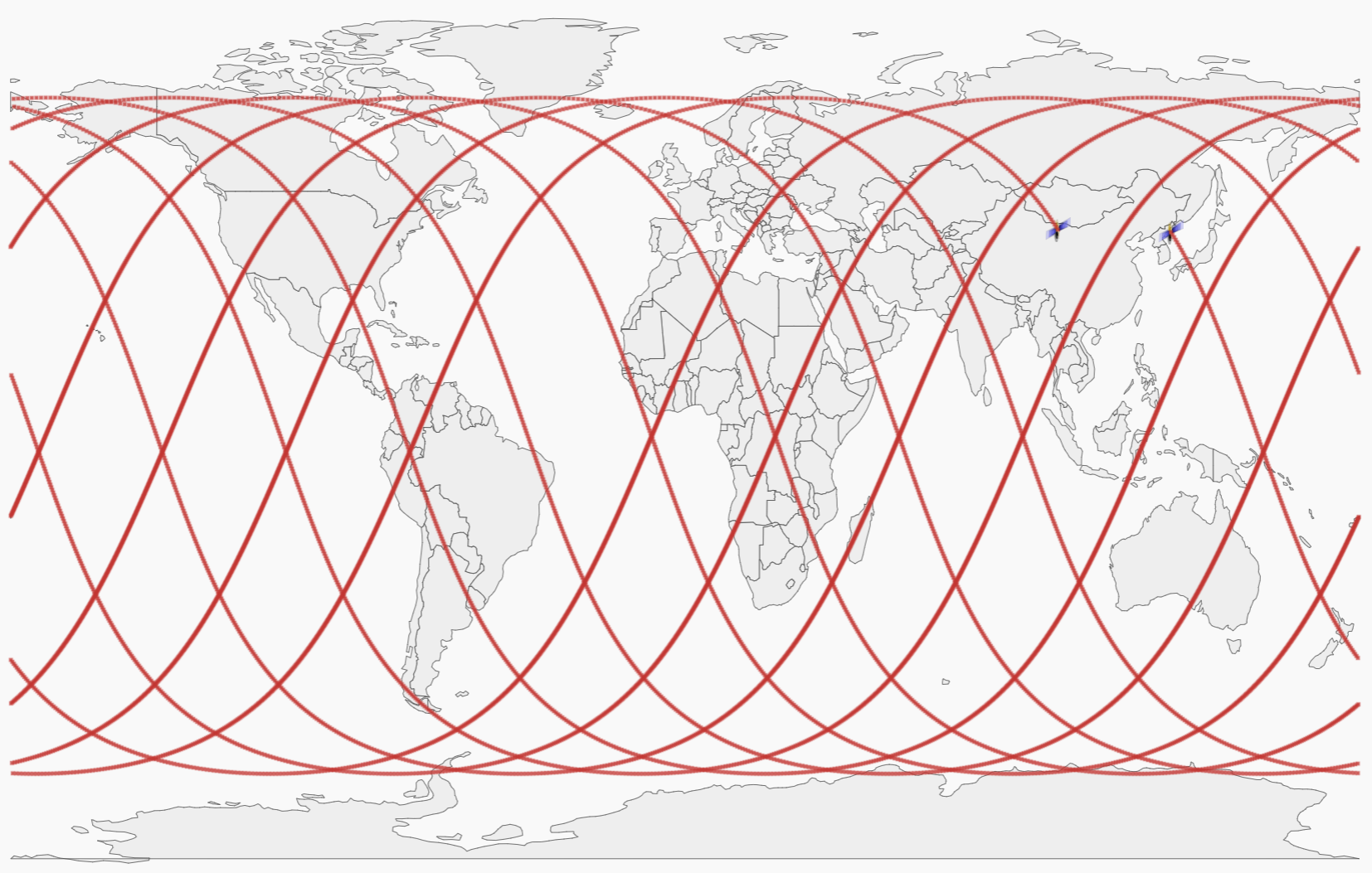}}
\caption{Calculated track of the subsatellite point}
\label{fig}
\end{figure}

\subsection{Satellite ground coverage calculation}
The coverage calculation is achieved by establishing the initial observation vector based on the geometry and parameters of the sensor field of view. At the same time, calculating the mapped positions on the Earth's surface corresponding to the sensor boundary units is based on the ephemeris parameters of the satellite, attitude parameters, and the Earth ellipsoidal model [4].

\vspace{1ex}
For all types of sensors, their coverage boundaries can be considered as the intersection of their sensor boundary point observation vectors with the ground, as shown in Figure 3.

\begin{figure}[htbp]
\centerline{\includegraphics[width=2in]{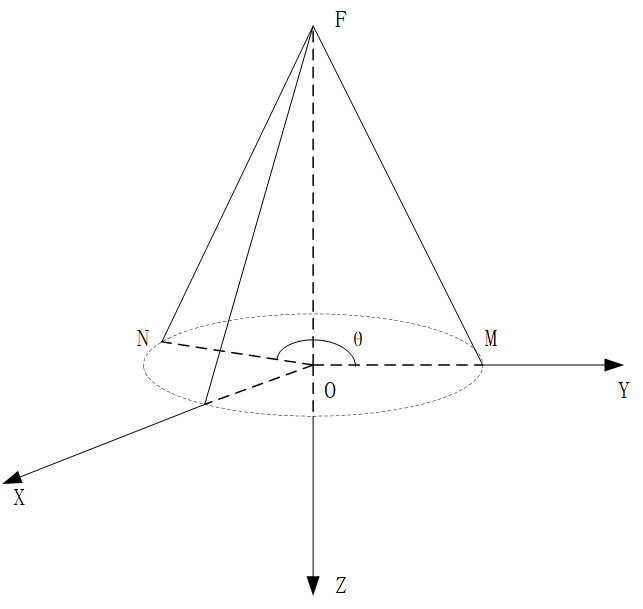}}
\caption{Schematic diagram of satellite imaging coverage area}
\label{fig}
\end{figure}

The sensor surface center is the origin O, the Z-axis points to the center of the earth, the X-axis is the satellite flight direction, the Y-axis is perpendicular to the flight direction, point N is any point on the sensor circle boundary, point M is a point on the boundary and on the Y-axis, then the observation vector $\overset{\rightarrow}{FN}$ at any point on the garden boundary can be obtained by turning $\overset{\rightarrow}{FM}$ along the Z-axis by an angle of $\theta$. Note that $\overset{\rightarrow}{l}$ is the observation vector of the sensor and $\overset{\rightarrow}{l}$ is $\left\lbrack {x,y,z} \right\rbrack^{T}$. Let $\left| {OF} \right| = f$, the conic section circle radius $\left| {OM} \right| = r$, then ${\overset{\rightarrow}{M} = \left\lbrack {0,r,0} \right\rbrack}^{T}$, $\overset{\rightarrow}{F} = \left\lbrack {0,0,\text{-}f} \right\rbrack^{T}$, $\overset{\rightarrow}{FM} = \left\lbrack {0,r,f} \right\rbrack^{T}$, and the specific definition is shown below.

\begin{equation}
\overset{\rightarrow}{l} = \begin{bmatrix}
{\mathit{\cos}\theta} & {\mathit{\sin}\theta} & 0 \\
{- {\mathit{\sin}\theta}} & {\mathit{\cos}\theta} & 0 \\
0 & 0 & 1 \\
\end{bmatrix}\overset{¯}{\text{FM}} = \begin{bmatrix}
{r{\mathit{\sin}\theta}} \\
{r{\mathit{\cos}\theta}} \\
f \\
\end{bmatrix}\label{eq}
\end{equation}

Let ${\frac{r}{f} = \mathit{\tan}}\alpha$, and then normalize to obtain

\begin{equation}
\left\{ \begin{matrix}
{x = {\mathit{\tan}\alpha}*{\mathit{\sin}\theta}} \\
{y = {\mathit{\tan}\alpha}*{\mathit{\cos}\theta}} \\
{z = 1} \\
\end{matrix} \right.\label{eq}
\end{equation}

Where $\alpha$ is the half-field-of-view angle and $\theta$ is the middle angle in the half-field-of-view range of the sensor. $\theta = \frac{2\pi}{n}$, and when $n$ is 2, it is a line-array push-and-sweep sensor. When $n$ is 4, it is a frame-amplitude sensor. 

\begin{equation}
\left\{ \begin{matrix}
{L = {\mathit{\arctan}\frac{Y}{X}}} \\
{B = {\mathit{\arctan}\frac{Z}{\sqrt{X^{2} + Y^{2}}}}\left( 1 + \frac{ae^{2}}{Z} \cdot \frac{\mathit{\sin}B}{\sqrt{1 - e^{2}{\mathit{\sin}^{2}B}}} \right)} \\
{H = \frac{\sqrt{X^{2} + Y^{2}}}{\mathit{\cos}B} - \frac{a}{\sqrt{1 - e^{2}{\mathit{\sin}^{2}B}}}} \\
\end{matrix} \right.\label{eq}
\end{equation}

Equation (9) is used to convert the spatial Cartesian coordinates into latitude, longitude, and elevation in the geodetic coordinate system. The resulting computed coverage simulation model of the frame-width sensor satellite is shown in Figure 4.

\begin{figure}[htbp]
\centerline{\includegraphics[width=2.3in]{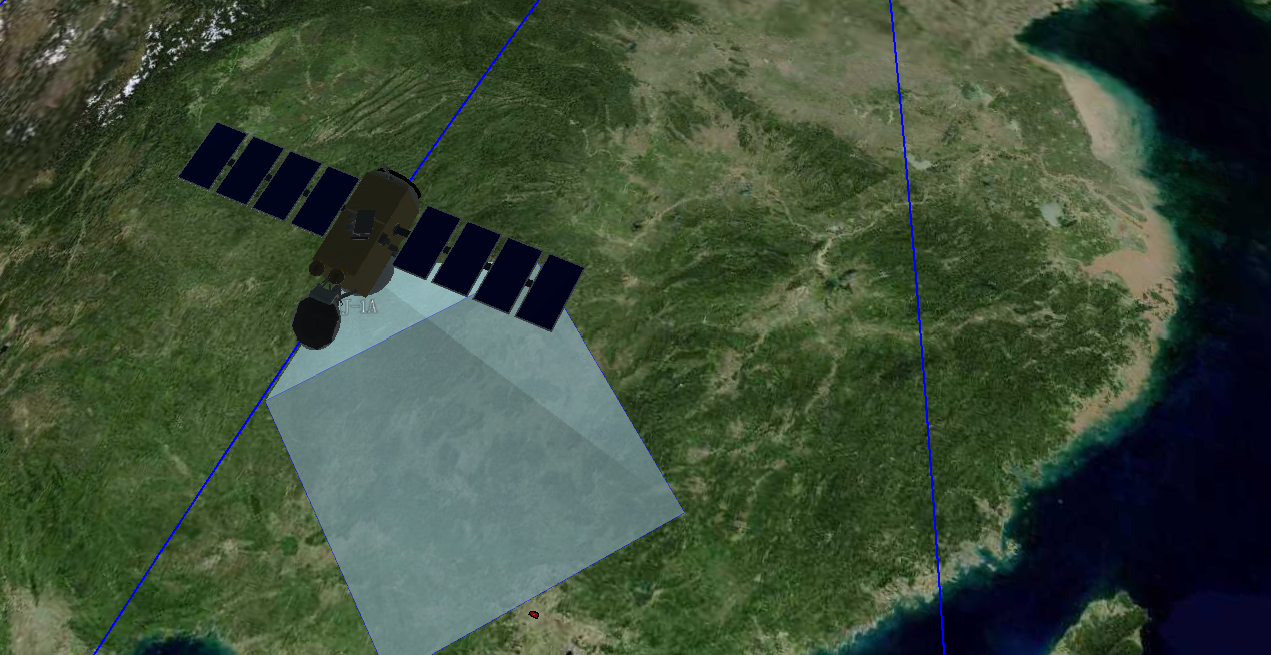}}
\caption{Simulation calculation of satellite ground coverage}
\label{fig}
\end{figure}

\section{Satellite state space model}
The Walker constellation configuration has good global and latitudinal band coverage characteristics and is widely used. All the satellite orbits in this constellation are nearly circular with equal semi-long axis, eccentricity, and orbital inclination, and the satellites are evenly distributed in the same orbital plane.

\vspace{1ex}
The configuration code of a Walker constellation is $N/P/F$ (number of satellites/number of orbital planes/phase factor). The longitude of ascending node and the argument of latitude of any satellite numbered m in the constellation are.

\begin{equation}
\Omega_{m} = \frac{360}{P}\left( {P_{m} - 1} \right)\label{eq}
\end{equation}

\begin{equation}
u_{m} = \frac{360}{S}\left( {N_{m} - 1} \right) + \frac{360}{N}F\left( {P_{m} - 1} \right)\label{eq}
\end{equation}

Where $S$ is the number of satellites in each orbital plane, $P_{m}$ is the number of the orbital plane in which the satellite is located, and $N_{m}$ is the number of the satellite in the orbital plane. That is, $S = \frac{N}{P}$,$P_{m} = \frac{m}{S} - 1$, $N_{m} = m - \left( {P_{m} - 1} \right)S$

\vspace{1ex}
To meet the coverage requirement of the target area with good global coverage performance, we choose the Walker constellation configuration as the experimental configuration in this paper to search for the optimal solution in this state space. The solutions for other constellations can also be calculated using the algorithm proposed below.

\section{Real-time coverage evaluation method with hexagonal discretization}
In order to quickly calculate the coverage of the target area by multiple satellites in a given time range [5], this paper proposes a new algorithm based on a dissected hexagonal grid: 
\begin{enumerate}
    \item Discretize the hexagon of the target area
    \item Determine the intersection of each grid with the satellite coverage zone
    \item Divide the number of intersecting grids by the number of all grids as the coverage at the current moment
    \item Obtain the average coverage by counting the coverage at different time periods
\end{enumerate}

\begin{figure}[htbp]
\centerline{\includegraphics[width=3in]{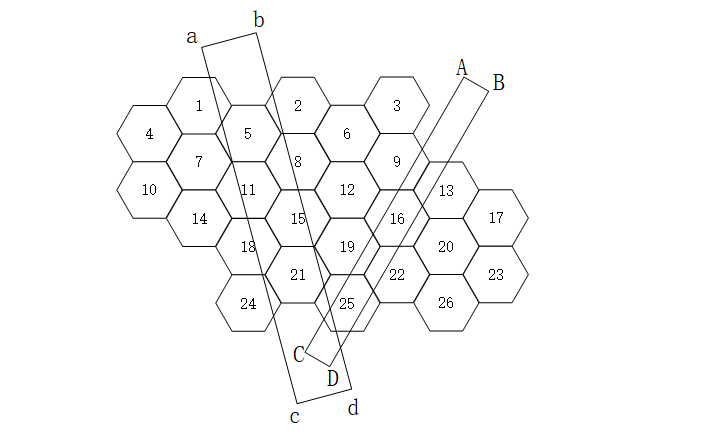}}
\caption{Coverage analysis method based on hexagonal grid}
\label{fig}
\end{figure}

As shown in Figure 5, there are 26 hexagonal grids in the target area, and the satellite coverage bands are $abcd$ and $ABCD$. The grids intersecting with $abcd$ are 1, 2, 5, 8, 11, 15, 18, 21, 24, 25, and the grids intersecting with $ABCD$ are 9, 13, 16, 19, 22, 25. 15 grids are covered, and the coverage rate is $\frac{15}{26} = 54.2\%$. 
 
\begin{figure}[htbp]
\centerline{\includegraphics[width=3in]{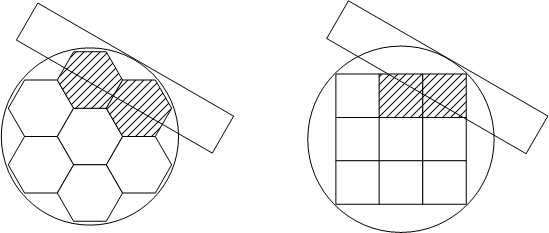}}
\caption{Coverage analysis method based on hexagonal grid}
\label{fig}
\end{figure}

As shown in Figure 6, the hexagonal grid method is closer to the actual covered area in the coverage analysis of circular areas than the meridional grid method. It has better performance in dealing with irregular areas.

\section{Constellation configuration overlay optimization design algorithm}
The simulated annealing algorithm is derived from the solid annealing principle. This probability-based algorithm heats the solid to a sufficiently high temperature and then lets it cool down slowly.

\vspace{1ex}
To ensure the uniformity of revisit time and meet the revisit time demand in different latitudes, a multi-inclination orbit combination scheme is used here, in which multiple satellites with the same inclination angle adopt the $Walker - \delta$ distribution [6]. For larger area targets with larger state space, the idea of a simulated annealing algorithm is considered to accelerate the search speed. When a satellite constellation configuration satisfying the revisit time requirement is not found, the algorithm will search it in a certain way in the range where the number of satellites increases; when a satellite constellation configuration satisfying the revisit time is found, the algorithm will search it in the range where the number of satellites is less than that satellite constellation configuration. The specific steps are as follows.

\vspace{1ex}
\begin{enumerate}
\item Given a higher initial temperature $T = T_{0}$, enter an initial solution
\item Determine whether the target region in this state satisfies the coverage constraint, if yes, go to step 3; otherwise, go to step 6
\item Store the solution as the better solution, and if the solution is the current solution with the least number of satellites used, then it will be the current optimal solution, otherwise with probability $exp\left\lbrack {- 100/\left\lbrack {n*T} \right\rbrack} \right\rbrack$ the solution is the current optimal solution, where $n$ is the number of iterations
\item Change the orbital inclination angle within a certain range according to the current optimal solution and coverage area, which decreases with temperature, and take the case of maximum average coverage
\item Subtract 1 from the number of orbital planes or the number of satellites in the plane corresponding to each inclination angle of the current optimal solution with a probability of $50\% T$. If the annealing is finished, go to step 7, otherwise go to step 2
\item Add 1 to the number of orbital planes or the number of satellites in the plane corresponding to each inclination angle of the current optimal solution with a probability of $50\% T$. If the annealing is finished, go to step 7, otherwise go to step 2
\item Output the current optimal solution as the optimal constellation configuration
\end{enumerate}

\vspace{1ex}
The algorithm steps are shown in Figure 7.

\begin{figure}[htbp]
\centerline{\includegraphics[width=3.3in]{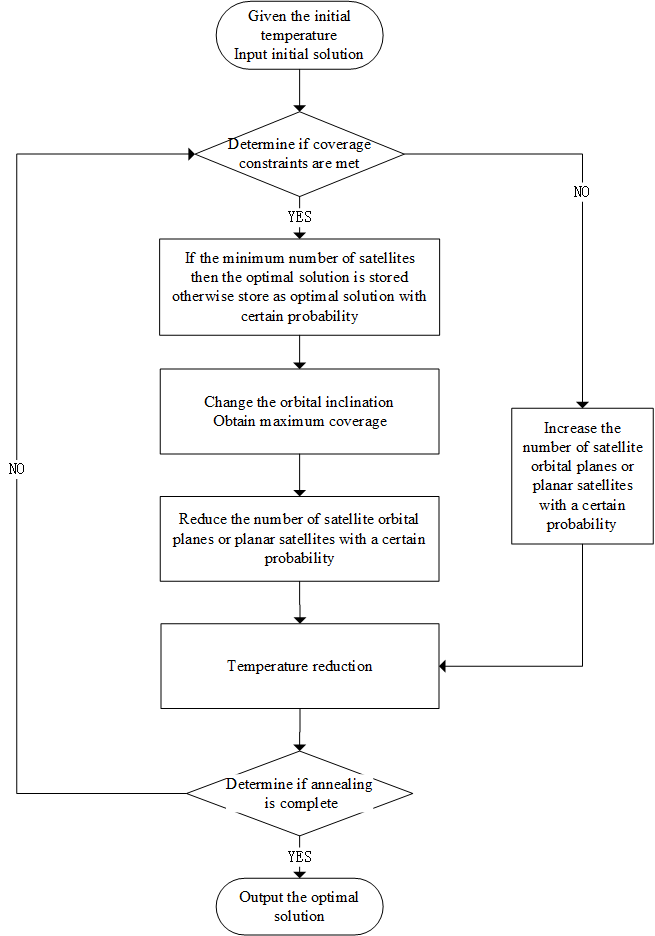}}
\caption{Coverage analysis method based on hexagonal grid}
\label{fig}
\end{figure}

\section{Simulation experiments}
\begin{table}[htbp]
\caption{Satellite experiment parameters}
\centering
\begin{tabular}{|c|c|}
\hline
Parameters                                                & Value                                                                                                                                                                                                                \\ \hline
orbital   radius                                          & 8576km                                                                                                                                                                                                               \\ \hline
eccentricity                                              & 0.0                                                                                                                                                                                                                    \\ \hline
true   anomal                                             & 0.0°                                                                                                                                                                                                                 \\ \hline
effective   field of view                                 & 60°                                                                                                                                                                                                                  \\ \hline
satellite   sensor type                                   & frame width sensors                                                                                                                                                                                                  \\ \hline
\multirow{4}{*}{Target coverage area boundary points} & \multirow{4}{*}{\begin{tabular}[c]{@{}c@{}}{[}100.382447° N, 19.47806° E{]}\\      {[}100.382447° N, 43.47806° E{]}\\      {[}124.382447° N, 43.47806° E{]}\\      {[}124.382447° N, 19.47806° E{]}\end{tabular}} \\
                                                          &                                                                                                                                                                                                                      \\
                                                          &                                                                                                                                                                                                                      \\
                                                          &                                                                                                                                                                                                                      \\ \hline
\end{tabular}
\end{table}

Experiments are conducted according to the satellite parameters described in Table 2. The experimental environment was Python 3.7 with 16G of computing memory, and the coverage calculation was hosted using ArcGIS API for Python to improve computing efficiency. The initial constellation is set to $i = 40^{\circ}\ Walker\ 6/3/1$, and the parameters of the simulated annealing algorithm are selected as $T = 1$, $T_{min} = 0.01$, $\alpha = 0.98$. The average coverage is defined as the average of the instantaneous coverage of any of the 50 selected moments during 30 satellite operation periods [7]. The minimum number of satellites is required to search the constellation for the region's average coverage of $70\%$. The optimal solution is obtained when the number of satellites is 54, the average coverage is $72.7\%$, and the configuration of the satellite constellation is $i = 38^{\circ}\ Walker\ 54/9/1$, where the Walker constellation configuration code is $N/P/F$ (number of satellites/number of orbital planes/phase factor). 

\vspace{1ex}
Figure 8 shows the instantaneous coverage of this optimized constellation for 100 moments. We can see that the constellation has a relatively uniform coverage of the target area with good performance[8].

\vspace{1ex}
\begin{figure}[htbp]
\centerline{\includegraphics[width=3in]{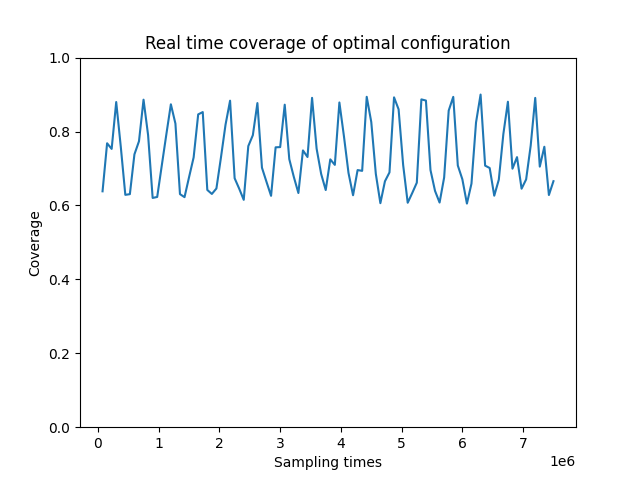}}
\caption{Variation of the instantaneous coverage of the best-configuration 
constellation during the operational cycle}
\label{fig}
\end{figure}

\begin{figure}[htbp]
\centerline{\includegraphics[width=3in]{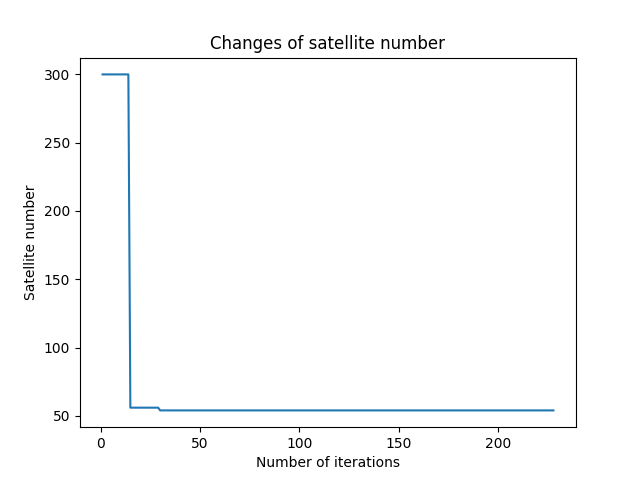}}
\caption{Variation of the number of optimal solution satellites with the number of iterations}
\label{fig}
\end{figure}

\begin{figure}[htbp]
\centerline{\includegraphics[width=3in]{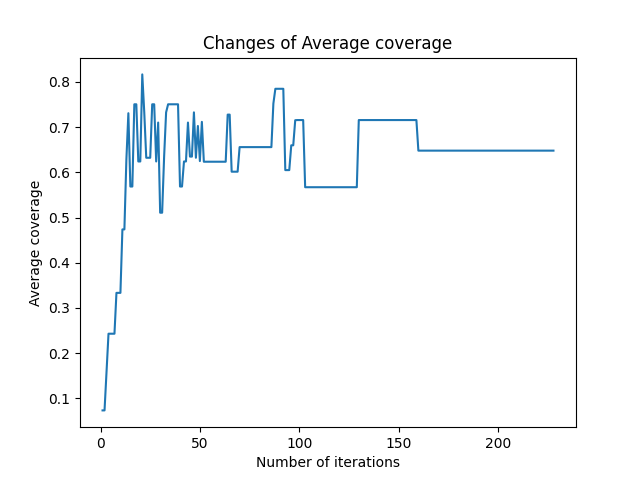}}
\caption{Variation of average coverage of constellations with the number of iterations during the search}
\label{fig}
\end{figure}

Figures 9 and 10 show the changes in obtaining the optimal solution and the changes in the average coverage of the constellation during the search. We can see that the improved simulated annealing algorithm can search for the optimal solution at the 26th iteration, which is faster, and the average coverage of the generated constellation is stable around the target range with the increase in the number of iterations, which has a good performance.

\section{Conclusion}
This paper proposes a constellation design and optimization method based on hexagonal discretization and a simulated annealing algorithm for large-scale imaging regions. Compared with traditional design methods, this method can reduce the number of satellites to be launched to complete the constellation construction task, thus reducing the time and cost required to complete the full satellite deployment. The reduction in the number of satellites also reduces the simultaneous co-location coverage area, thus reducing the amount of data to be processed and the processing time. The designed algorithm is scalable and scalable and can be modified to output multiple constellation configurations that meet the revisit time requirements. The solution can design and select constellation configurations faster, which is of great significance for area-specific surveillance, disaster monitoring, and military reconnaissance.

\section{Acknowledgment}
The authors would like to thank the anonymous reviewers for their valuable feedback and suggestions. The authors would also like to express their gratitude to their colleagues and mentors who provided support and guidance throughout Jiahao's writing process. Boheng and Jiahao made equal technical contribution for this work.

\end{document}